\documentclass[draft]{agujournal2019}
\usepackage{url} 
\usepackage{lineno}
\usepackage{amsmath}
\usepackage[inline]{trackchanges} 
\usepackage{soul}
\draftfalse

\journalname{Journal of Advances in Modeling Earth Systems (JAMES) }

\begin{document}

%
%


\title{Generative Data Assimilation of Sparse Weather Station Observations at Kilometer Scales}

%
%


\authors{Peter Manshausen\affil{1, 2}, Yair Cohen\affil{1}, Peter Harrington\affil{1},  Jaideep Pathak\affil{1}, Mike Pritchard\affil{1,3}, Piyush Garg\affil{1}, Morteza Mardani\affil{1}, Karthik Kashinath\affil{1},  Simon Byrne\affil{1}, Noah Brenowitz\affil{1}}

\affiliation{1}{NVIDIA, Santa Clara, CA, USA}
\affiliation{2}{University of Oxford, Oxford, UK}
\affiliation{3}{University of California Irvine, Irvine, CA, USA}


\correspondingauthor{Peter Manshausen}{peter.manshausen@physics.ox.ac.uk}



\begin{keypoints}
\item We demonstrate data assimilation of weather station data to 3km-resolution surface fields with a diffusion model surrogate.
\item This opens up a simple, scalable pipeline to create km-scale ensemble reanalyses at low cost and latency, competitive to operational ones.
\item The model is easily adapted to new observations, produces states of variables not directly observed, and shows evidence of learned physics.
\end{keypoints}

\justifying

\begin{abstract}
Data assimilation of observations into full atmospheric states is essential for weather forecast model initialization. Recently, methods for deep generative data assimilation have been proposed which allow for using new input data without retraining the model. They could also dramatically accelerate the costly data assimilation process used in operational regional weather models. Here, in a central US testbed, we demonstrate the viability of score-based data assimilation in the context of realistically complex km-scale weather. We train an unconditional diffusion model to generate snapshots of a state-of-the-art km-scale analysis product, the High Resolution Rapid Refresh. Then, using score-based data assimilation to incorporate sparse weather station data, the model produces maps of precipitation and surface winds. The generated fields display physically plausible structures, such as gust fronts, and sensitivity tests confirm learnt physics through multivariate relationships. Preliminary skill analysis shows the approach already outperforms a naive baseline of the High-Resolution Rapid Refresh system itself. By incorporating observations from 40 weather stations, 10\% lower RMSEs on left-out stations are attained. Despite some lingering imperfections such as insufficiently disperse ensemble DA estimates, we find the results overall an encouraging proof of concept, and the first at km-scale. It is a ripe time to explore extensions that combine increasingly ambitious regional state generators with an increasing set of in situ, ground-based, and satellite remote sensing data streams. 
\end{abstract}

\section*{Plain Language Summary}
Weather forecasts rely on our knowledge of the full state of the atmosphere in the present. Weather stations provide measurements at some (sparse) locations.The atmospheric variables need to be filled in with models that transform the point measurements to a full state (a map). Usually, such models are numerical weather models encoding the physical laws of the atmosphere. They are expensive and slow to run, limiting real-time updates to forecasts. However, recently the machine learning community has presented great advances in similar tasks like filling in missing parts of photographs, and even generating entire videos from a few words. This motivates our use of an ML model trained on km-scale weather data and guided by sparse point measurements to fill in wind and rain maps on the same scale (3km) as state-of-the-art conventional models.

\section{Introduction}
Machine Learning (ML) has attracted intense interest in the field of global weather forecasting, with models trained off reanalysis outperforming state-of-the-art numerical models \cite{keisler2022forecasting, pathak2022fourcastnet,bi2022pangu, lam2022graphcast}. More recent developments include skillful ensemble forecasting \cite{price2023gencast} and the integration of a numerical dynamical core with online ML parameterizations in one global circulation model \cite{kochkov2023neural}. In order to forecast weather at km-scale, \citeA{mardani2023residual} propose downscaling coarse resolution forecasts with diffusion models \cite{karras2022elucidating}. Together, these advances can be viewed as a disruption of traditional physics-based weather prediction using data-driven approaches stemming from the image and video computer science research community. Dynamical tests suggest that ML video generation is a paradigm that has captured physically realistic responses inherent in weather prediction models \cite{hakim2024dynamical}. 

It is natural to wonder if similar transdisciplinary disruptions will modify how the weather and climate communities approach the separate task of \textit{state estimation}. Forecast initialization and many other applications such as nowcasting depend on high-resolution data assimilation: combining the latest (sparse) observations with our knowledge of the physical laws governing the atmosphere. This inference problem of estimating a full, dense atmospheric state from observations is traditionally done by constraining numerical models with the available observations. Methods for this include Kalman filters \cite{anderson2001ensemble} and variational methods, see \citeA{bannister2017review} for a review.
Similar inference problems, like inpainting gaps in photographs, and imagining entirely new parts of images, have been successfully addressed with diffusion models \cite{rombach2022highresolution,lugmayr2022repaint}. 

Data assimilation systems for regional domains can be quite complicated.
For example, a premier US regional model---the high resolution rapid refresh (HRRR)\cite{dowell2022high}---depends on outputs of two other data assimilation systems, the Global Forecast System (GFS) and the Rapid Refresh (RAP) \cite{Benjamin2016-ma}. Not all fields and observations are treated consistently, as the dynamical fields (winds, pressure) use classic ensemble methods, while the radar is turned into a latent heating which directly updates the model's thermodynamics.
A separate digital filter is required to avoid high-frequency spin-up artifacts when the observations clash with the model's fast processes by e.g. violating its notion of hydrostatic balance or saturation \cite{Lynch2003-ju}. Many of these separate steps are addressed in a single step by the ECMWF data assimilation scheme \cite{rabier2000ecmwf} for the global domain, but the data assimilation process remains computationally expensive (at least a 1h delay between the last observations assimilated and the start of forecasting). Thus, there is an opportunity to simplify and improve the accuracy of these complex pipelines by training cheap ML emulators which admit a simpler formulation of data assimilation algorithms. While an ML-based data assimilation approach would not necessarily solve the issue of high-frequency spin-up artifacts in numerical forecasting models, ML forecasting models—such as StormCast \cite{pathak2024kilometer} for the CONUS domain—seem less sensitive to these \cite{hakim2024dynamical}. 

Several recent studies already suggest the potential of training a diffusion model emulator of a numerical model-- \citeA{rozet2023scorebased} propose Score-based Data Assimilation (SDA), where a similar approach is used to reconstruct states of dynamical systems from observations, experimenting on systems as complex as the two-layer quasigeostrophic equations \cite{rozet2023scorebasedqg}. This framework has been used in the global weather context, assimilating weather stations and satellite data for 2m temperature at 25km resolution \cite{qu2024deep}. In related work, \citeA{huang2024diffda} assimilate pseudo-observations of reanalysis data in an autoregressive manner similar to operational data assimilation. These approaches do not include observations in the training of ML models, but only analysis data. The observations are only used in the inference phase. Therefore, the models can be used flexibly to assimilate new observational data streams. This is in contrast to the approach of \citeA{andrychowicz2023deep}. Here, the authors train with pairs of observation data and numerical weather model assimilated data from different times. Their model takes such a pair as input and outputs a forecast of `densified' station measurements. This means that, to add a new source of observational data, the model needs to be retrained.   

Open questions remain about the performance and calibration of these approaches, and about whether models encode physical relationships. Furthermore, for many impact variables like storms and precipitation, km-scale resolutions are useful. Here, we apply the SDA framework to 3km resolution data of wind and precipitation in the central US. We train a diffusion model with analysis data, and show assimilation of weather station data of the same variables. Section~\ref{sec:methods} presents the diffusion model and the SDA framework, section~\ref{sec:data} discusses the weather station and analysis data used, section~\ref{sec:results} shows the assimilation of data in different settings, and section~\ref{sec:conclusions} discusses the results.

\section{Materials and Methods}\label{sec:methods}
In this work, we adopt the framework of Score-based Data Assimilation (SDA) by \citeA{rozet2023scorebased} with only minor adaptations. The goal of data assimilation here is to find the posterior of the system state $\textbf{x}$ given observations $\mathbf{y}$, $p(\mathbf{x} |\mathbf{ y})$. Instead of using an numerical weather model to generate or update system states $\mathbf{x}$, we train a diffusion model for this task. We briefly present diffusion models, then SDA, and then our own implementation. Note, that in this description, we adopt the unified notation for data assimilation after \citeA{ide1997unified}, rather than the notation used by \citeA{rozet2023scorebased}. We list the equivalent notations in \ref{app:notation}.
\begin{figure}
\centering
\noindent\includegraphics[width=\textwidth]{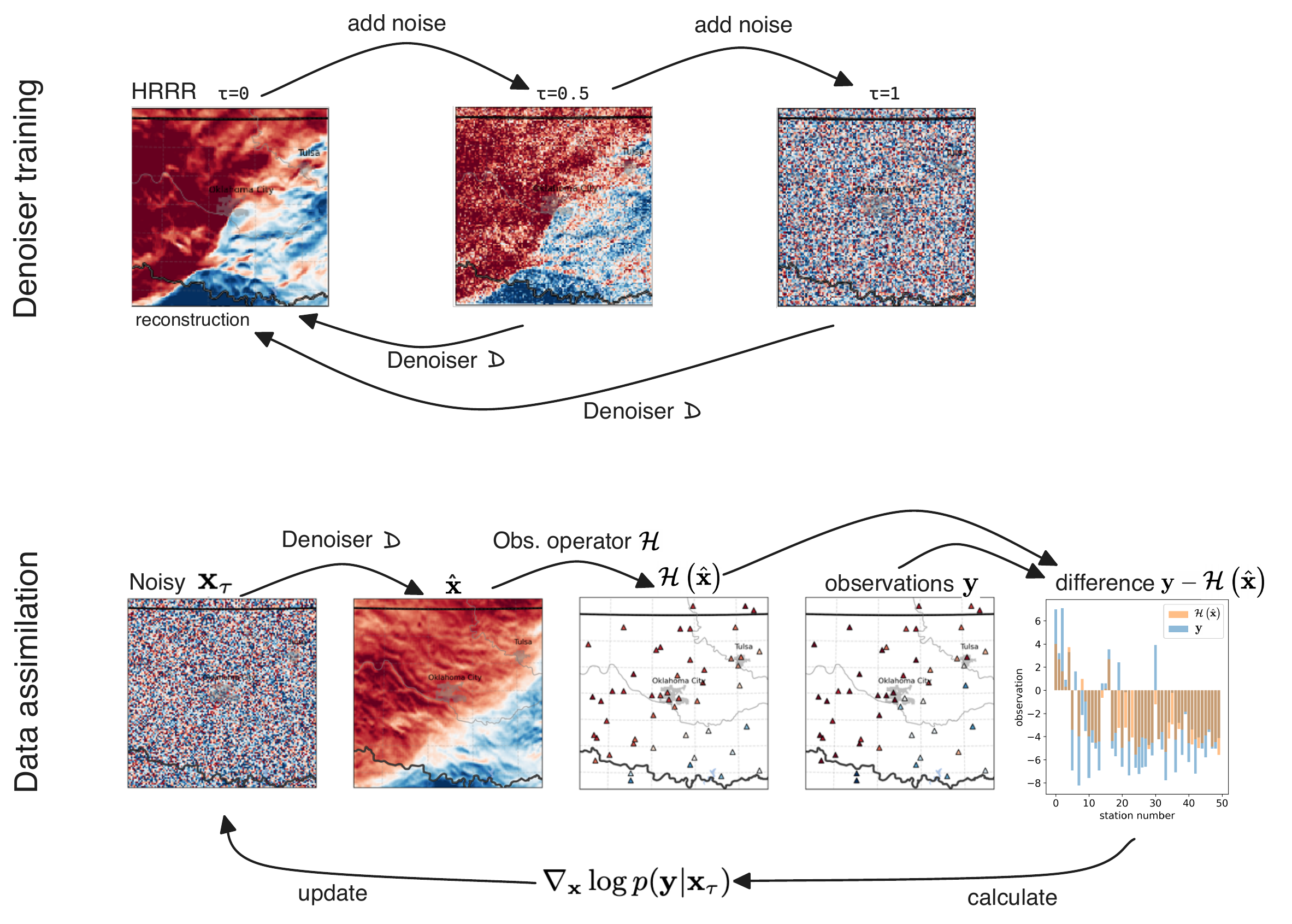}
\caption{Denoiser training and data assimilation with SDA. \textbf{a)} During the training of the denoiser, noise is added to the training data at different levels, parameterized by diffusion-time $\tau\in [0,1]$. The training objective for the denoiser $D$ is to reconstruct the training data, given the noisy state and the time $\tau$. \textbf{b)} Data assimilation then uses $D$ to go from a noisy state $\mathbf{x}_\tau$ to a possible denoised state $\hat{\mathbf{x}}$. The observation operator $\mathcal{H}$ then maps $\hat{\mathbf{x}}$ to the observations it would give rise to, which we compare to the actual observations $\mathbf{y}$ (e.g. weather station data). The difference of the two is used to calculate the score $\nabla_{\mathbf{x}} \log p(\mathbf{y}|\mathbf{x}_\tau)$, giving the direction in which the noisy state $\mathbf{x}_\tau$ is updated. This cycle repeats for a number of steps, with time running from 1 to 0, until $\mathbf{x}$ is denoised, taking into account the observations and the model's learned prior (the HRRR reanalysis).}
\label{fig:method}
\end{figure}

\subsection{Diffusion Models}
Score-based data assimilation builds on the score-matching formulation of denoising diffusion models \cite{song2019generative, ho2020denoising}.  Score-based generative models, or diffusion models, are a type of machine learning model that can generate samples from highly complex data distributions. They consist of a forward process, a reverse process, and a sampling procedure. 

Briefly, a \emph{forward} process maps the data---$\mathbf{x}_0$--onto a known noise distribution $\mathbf{x}_T$, with pseudo-time $\tau$ running from 0 to $T$. The distribution then follows
\begin{equation}
    p(\mathbf{x}_\tau |\mathbf{x}_0) = \mathcal{N}(\mathbf{x}_\tau |\mu_\tau \mathbf{x}_0,\Sigma_\tau)
\end{equation}
with $\Sigma = \sigma_\tau I$  and both $\sigma_\tau$ and  $\mu_\tau$ are parameterizations we can choose. In our case, they  are both simply given by $\sigma_\tau=\tau$ and $\mu_\tau = 1$  yielding the simple evolution $\mathbf{x}_\tau=\mathbf{x}_0 + \tau \epsilon$ where $\epsilon$ is Gaussian white noise \cite{karras2022elucidating}. Intuitively, we can always make data look like noise by adding a lot of noise.
It turns out---nearly by magic---that this noising process can be reversed (the \textit{reverse} process) if one knows the so-called \emph{score-function} $\nabla_{\mathbf{x}} \log(p(\mathbf{x}))$. In practice large neural networks can be trained to approximate the score-function. This can be done by providing the network with samples of noisy images $\mathbf{x}_\tau$, the noise level given by pseudo-time $\tau$, and the denoised image $\mathbf{x}_0$. Figure~\ref{fig:method}a) shows the objective which the model, the denoiser $D$, is trained with: Given a sample from the training data with noise added at a certain amplitude, the model is trained to reconstruct the original sample---a process called denoising. Given the trained denoiser, new samples from the distribution $p(\mathbf{x})$ can be drawn with the \textit{sampling procedure}:  Given a sample of Gaussian white noise, this is taken to be $\mathbf{x}_T$. We divide $T$ into $N$ timesteps, and perform denoising steps $d\mathbf{x} \propto \mathbf{x}_\tau- D( \mathbf{x_\tau}, \tau) $ in the direction of the denoised state, thereby discretizing the reverse process. 
See \citeA[algorithm 2]{karras2022elucidating} for an implementation of the sampling procedure. Being able to sample from the prior distribution $p(\mathbf{x})$ will help our goal of sampling from the posterior $p(\mathbf{x} |\mathbf{ y})$.

\subsection{Score-based Data Assimilation}

The score-based formulation is so useful for data assimilation because it permits an elegant formulation of Bayes' rule,
\begin{equation}
        \nabla_{\mathbf{x}} \log p(\mathbf{x}_\tau|\mathbf{y}) =  \nabla_{\mathbf{x}} \log p(\mathbf{x}_\tau) +  \nabla_{\mathbf{x}} \log p(\mathbf{y} | \mathbf{x}_\tau) 
\end{equation}
where $0\leq \tau <T$ is the ``time'' of the noising process, not real-world time.
The first term---the score function---is given by the learned denoiser $D$: This is the model described in the previous section, which can generate dynamic system states from noise. In our case, these system states are snapshots of atmospheric states which follow the same distribution as (``look like") the training data. The second term is related to the likelihood: Other data assimilation techniques can be derived using a similar maximum likelihood approach that maximizes $ p(\mathbf{y} | \mathbf{x})$ . This maximization is done by taking its gradient and finding its zeros, or equivalently those of the gradient of the logarithm, as the logarithm is monotonic—see below for a comparison to 3DVar. 

Unfortunately, in the diffusion-based case $ p(\mathbf{y} | \mathbf{x}_\tau) $ is only known at $\tau=0$---recall that $\tau=0$ corresponds to the data distribution. If the probability distribution of the denoised state given the noisy one $p(\mathbf{x}_0|\mathbf{x}_\tau)$ has non-zero variance, then  $ p(\mathbf{y} | \mathbf{x}_\tau) $ is different from $ p(\mathbf{y} | \mathbf{x}_0) $.

\citeA{rozet2023scorebased} follow \citeA{chung2022diffusion} in assuming a Gaussian observation process, but add a noise term to the variance to account for the approximation at $\tau \neq 0 $. The full likelihood in SDA is then
\begin{equation}
    p(\mathbf{y}| \mathbf{x}_\tau) = \mathcal{N}\big(\mathbf{y} | \mathcal{H}\left(\hat{\mathbf{x}}\right), \mathbf{R} + \frac{\sigma_\tau^2}{\mu_\tau^2}\Gamma\big).
    \label{eq:sda-likelihood}
\end{equation}

This means that the probability of observations $\mathbf{y}$ given the noisy state $\mathbf{x}_\tau$ is a normal distribution around a mean given by $\mathcal{H}(\hat{\mathbf{x}})$, which describes the differentiable observation operator $\mathcal{H}$ acting on the denoised state $\hat{\mathbf{x}}=D(\mathbf{x}_\tau; \tau))$. The observation operator maps from the state to the observations (in our case, $\mathcal{H}$ just selects the locations where we have observations from the full state). The variance of the distribution is given by two terms: The noise of the observations $\mathbf{R}$, and the term parameterized by $\Gamma$, which accounts for the approximation of using the denoised $\hat{\mathbf{x}}$. In this term, $\sigma$ and $\mu$ parameterize the noising process of the diffusion model, and $\Gamma$ is a hyperparameter we can choose (\citeA{rozet2023scorebased} choose a single-valued diagonal matrix, and we follow this choice). Overall, the Gaussian assumption for modeling the effect of using the denoised $\hat{\mathbf{x}}$ instead of $\mathbf{x}_\tau$ in the likelihood score could impact the spread of the resulting samples since it is mode-seeking. It replaces a potentially complex distribution with a Gaussian distribution. To stop errors from accumulating during the sampling process, following \citeA{rozet2023scorebased} (Algorithm 4), we perform $C$ steps of Langevin Monte Carlo (LMC), adding noise and denoising with step size $\delta$.

In the inference phase, the posterior score is used for the generation of a full state—this is called `guidance' \cite{rombach2022highresolution, mardani2023variational, ho2022classifier}. Figure~\ref{fig:method}(b) shows how the observations are used for guidance in the denoising process. Using the same sampling procedure as for unguided generation, the only change comes from using the posterior score instead of the prior score: Given the noisy state $\mathbf{x}(\tau)$, the likelihood score is calculated by denoising using $D$, applying the observation operator $\mathcal{H}$, calculating the squared difference with the observations $\mathbf{y}$, dividing by the variance, and taking the gradient. The state is then updated in the direction of the sum of the gradients of prior and likelihood. This process is repeated for $N$ steps, as before in the sampling of the reverse process. 

We obtain the score by taking the logarithm of the likelihood, so equation~\ref{eq:sda-likelihood} takes the shape
\begin{equation}
    \log p(\mathbf{y}| \mathbf{x}_\tau) = -\frac{1}{2}(\mathbf{y} - \mathcal{H} (\hat{\mathbf{x}}))^T \mathbf{V}^{-1} (\mathbf{y} - \mathcal{H} (\hat{\mathbf{x}})).
\end{equation}
where $\mathbf{V} = (\mathbf{R} + \frac{\sigma_\tau^2}{\mu_\tau^2}\Gamma)$. This takes very similar shape as the second term of the cost function that 3D-Var \cite{courtier1998ecmwf} is seeking to minimize with respect to $\mathbf{x}$, given by 
\begin{equation}
J = (\mathbf{x}_b - \mathbf{x})^T \mathbf{B}^{-1} (\mathbf{x}_b - \mathbf{x}) + (\mathbf{y} - \mathcal{H}(\mathbf{x}))^T \mathbf{R}^{-1} (\mathbf{y} - \mathcal{H}(\mathbf{x}))\label{eq:3dvar}
\end{equation}
where $\mathbf{x}_b$ is the background state, $\mathbf{B}$ is the covariance matrix of the background error. This makes sense, as the second term in the cost function gives the `update' resulting from the observations. The first term in eq.~\ref{eq:3dvar} can be thought of as the 3D-Var analogue of the prior $\log p(\mathbf{x})$, where 3D-Var typically starts from a previous forecast for $\mathbf{x}_b$. For a discussion of how  $ \left(\mathbf{y}-\mathcal{H}\left(\hat{\mathbf{x}}\right) \right)$ is used in a Kalman filter, see e.g. \citeA[eq. 9]{houtekamer2005ensemble}.

Crucially, the denoising model $D$ is not trained on the sparse observation data, but on full atmospheric states from analysis alone. Observations are only supplied at inference time, making the assimilation a `zero-shot' problem, i.e. the model was not explicitly trained with the kind of data it is used for. This means we do not need to retrain the denoiser when we want to incorporate new observations constraining the model's existing channels.

\subsection{Implementation of SDA for Kilometer-scale Weather}
We train a diffusion model to generate atmospheric states of 10m winds and surface precipitation in our study region in the central US. We are using the Elucidated Diffusion Models (EDM) framework of \citeA{karras2022elucidating}, which was also used in the CorrDiff downscaling model of \citeA{mardani2023residual}. Unlike CorrDiff, we do not condition the model on large-scale meteorology, but train an unconditional diffusion model which generates plausible snapshots of atmospheric states (no forecasting or time dimension). The model is trained to nearly 2.5M images of the 10m winds and surface precipitation from the HRRR analysis (see section~\ref{sec:data}). Training takes under 8 hours on 16 NVIDIA A100 (40GB) GPUs. An inference takes 10 seconds on a single NVIDIA RTX 6000 Ada Generation GPU  with the hyperparameters for the number of diffusion steps $N = 64$ and corrections $C = 2$ used for almost all experiments here (see \ref{app:hyperparams}).

Different from the SDA approaches of \citeA{rozet2023scorebased}, we do not use multiple (physical) time steps here. They train a diffusion model that generates a whole sequence of states $\mathbf{x}_t$, with $t = 1, ..., T $. In our case, we generate only individual system states, and assimilate only single time step observations (similar to 3D-Var). Other hyperparameters remain largely unchanged from their experiments, with an overview given in \ref{app:hyperparams}. \citeA{rozet2023scorebased} train their model with a slightly different objective from our denoiser, but the two approaches are compatible. A derivation of how to use a denoiser $D$ trained in the EDM framework of \citeA{karras2022elucidating} in the SDA framework is given in \ref{app:edmsda}. 

We include an evaluation of the unconditional generation in Appendix~\ref{app:uncond}. We find that while the means of the distributions are well represented, the distribution of generated states has lower variance than the training distribution. This is somewhat alleviated by guiding with observations in SDA. 

\section{Data}\label{sec:data}
For simplicity, we focus on a region roughly the size of a US state. The square region of interest is bounded by 37.197\textdegree ~to 33.738\textdegree N, 99.640\textdegree ~to 95.393\textdegree W, symmetric around Oklahoma City and covering most of the state of Oklahoma. This region is chosen mainly due to the stochastic nature of convective precipitation here and the density of the observational network, a setting that makes a stochastic method of data assimilation a natural choice. 

\subsection{High-Resolution Rapid Refresh (HRRR)}
The HRRR \cite{dowell2022high} is an hourly-updated, convection-allowing atmospheric model. It is an implementation of the Advanced Research version of the Weather Research and Forecasting (WRF-ARW) Model \cite{skamarockdescription} in the contiguous US and Alaska. It runs on a 3km grid, and assimilates a large variety of observational data. The coarser-resolution Rapid Refresh (RAP, \citeA{Benjamin2016-ma}) assimilates observations from aircraft, radiosondes, GPS precipitable water, METAR and mesonet stations, buoys/ships, profiler, and satellite winds) to a 13km grid. HRRR differs from this in its 3km resolution, and by assimilating three-dimensional radar-reflectivity data from NOAA's MRMS product \cite{zhang2016multi}.
From the full data, we extract the grid-native wind components u10m (easting) and v10m (northing).
While these do not correspond exactly with zonal and meridional winds, we will assume for simplicity that they do and refer to them accordingly in the rest of the manuscript.
Since the region considered is in the center of the HRRR domain, this a wind-direction error of up to $1.4^{\circ}$, which is likely smaller than the precision of the observations and other sources of error in our pipeline.
We also extract total surface precipitation over one hour (tp).
The values for 10u and 10v can be negative (blowing east-to-west and north-to-south, respectively).
This spatial subsetting gives patches of $128 \times 128$ pixels. We train with data spanning the period from 2018--2021 inclusive, validating the model training and tuning hyperparameters on 2022. We test on 2017 data. Updates of the HRRR methodology were found to cause nonstationarities in the HRRR data prior to 2018, affecting upper levels, but not the surface channels we use here.

\subsection{NOAA Integrated Surface Database (ISD)}\label{sec:ISD}
The ISD \cite{noaaisd} is a global, hourly dataset comprising more than 14,000 weather stations. Here, we download wind and precipitation data in our region of interest for the year of 2017 (out of the HRRR training sample). Wind speed and direction are turned into zonal and meridional surface velocity components. The weather station data is not reported at uniform times, so we interpolate in time to full hours. Our region spans 50 weather stations in the ISD, which in turn have data available in around a third (for wind) and a fifth (for precipitation) of times. It is important to note, that the stations we use are all part of the METAR wind data assimilated by HRRR (the precipitation data is not assimilated by HRRR). 

We need to model the observation error (noise) statistics in accordance with equation~\ref{eq:sda-likelihood} in order to weight the observations in the assimilation step against the model-learned prior. Following \citeA{rozet2023scorebased} we assume the covariance matrix $\mathbf{R}$ to be diagonal, i.e. for observation noise to be uncorrelated. The diagonal entries are taken to be the the same in each individual variable, and noted $\Sigma_{y} \in (\Sigma_P, \Sigma_u, \Sigma_v)$. For precipitation,
\[
    \log(P_i + 10^{-4}) | \mathbf{x} \sim  \mathcal{N}(\log(P_i^\mathrm{HRRR} + 10^{-4}), \Sigma_P)
\]
where $P_i^\mathrm{HRRR}$ is the precipitation of the nearest HRRR grid point, and $\Sigma_P$ is the estimated  observation process noise.

For the winds, 
\[
    \mathbf{u}_i | \mathbf{x} \sim  \mathcal{N}(\mathbf{u}_i^\mathrm{HRRR}, \Sigma_{\mathbf{u}})
\]
with $\mathbf{u}_i^\mathrm{HRRR}$ the two wind components at the nearest HRRR grid point. In the experiments, we find optimal values of $\Sigma_P$ and $\Sigma_{\mathbf{u}}$ by tuning for performance on left-out stations. For more details see \ref{app:hyperparams}.

\section{Results}\label{sec:results}
\subsection{Data Assimilation of Analysis Data}
\begin{figure}
\noindent\includegraphics[width=0.97\textwidth]{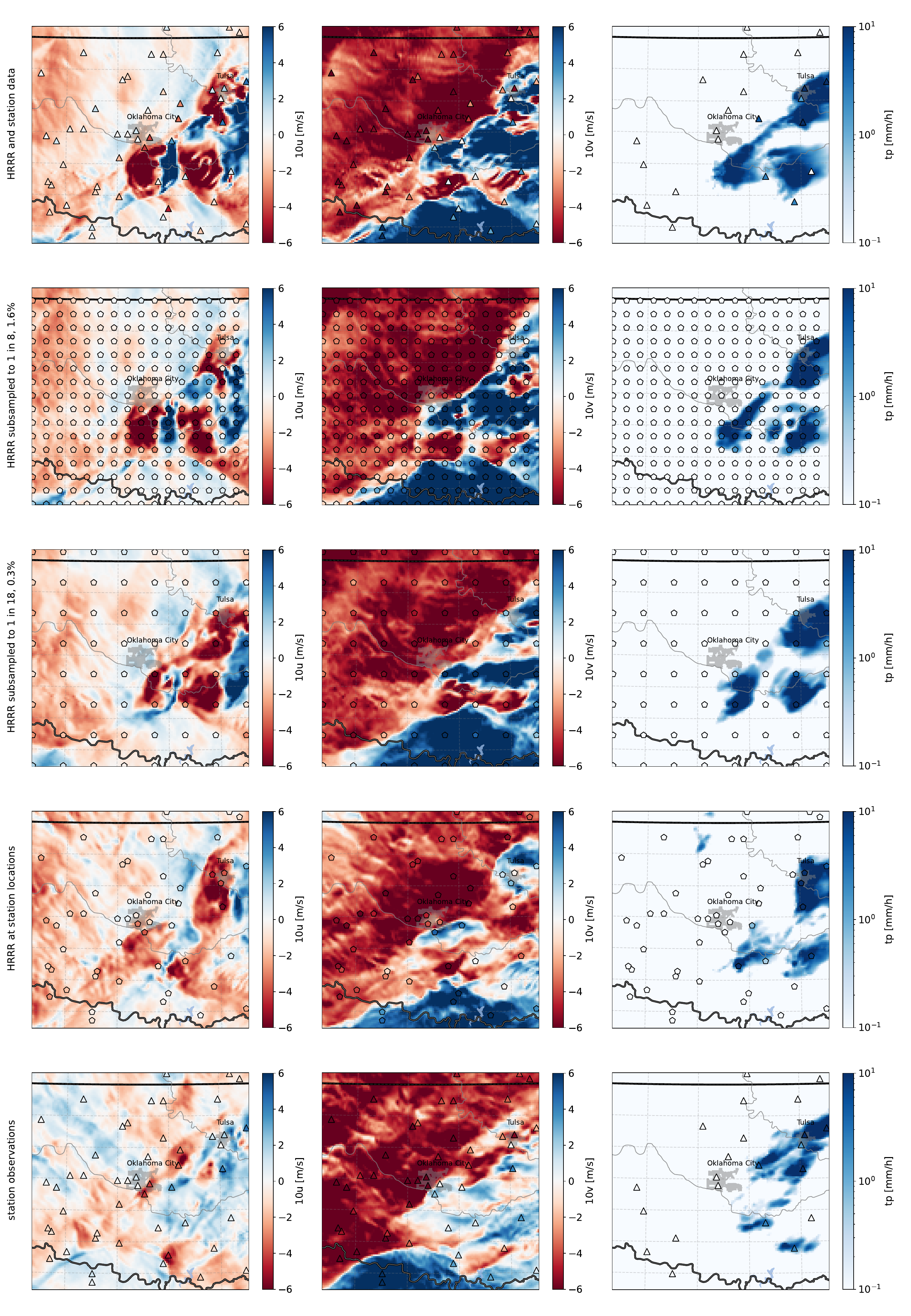}
\caption{Assimilating increasingly sparse and noisy data. Columns show the different variables 10u, 10v, and tp for different study cases. In row one, we show HRRR data of 2017-05-28 03:00 UTC, as well as the station data (triangles). Rows two and three show this data subsampled to 1.6\% and 0.3\%, respectively, in a regular grid (pentagons), plotted over the assimilated high-resolution state. Row four shows the HRRR data subsampled to the locations of our ISD weather stations, as well as the assimilated state. Row five shows again the observations from the stations (triangles), as well as the assimilated state. For a visualization of the HRRR winds as vector arrows, see Fig.~\ref{missingch}}
\label{subsample}
\end{figure}

We first demonstrate and build confidence in the method's validity in the limit of realistically sparse weather station data, by assimilating increasingly difficult (pseudo-) observational data to an atmospheric state in Fig.~\ref{subsample}. For the pseudo-observations, we use HRRR data from the year of 2017, which was not part of the training data set. We subsample the HRRR data to be increasingly sparse, and perform data assimilation. This way, we can show how SDA performs when the true full state corresponding to sparse data is known. This is not the case when we assimilate real weather station data; HRRR is not necessarily representative of the true atmospheric state that produced the observations, as can be seen by the mismatch of HRRR and station data in the first row. The SDA algorithm assumes the pseudo-observations to have the same observation error statistics as the real data, modeled in section~\ref{sec:ISD}. We start with a snapshot of HRRR data in the top row, displaying the two wind variables, and precipitation, which the model is trained for. We first subsample the data to a regular grid of one in eight latitude and longitude positions, resulting in a fraction of $(\frac{1}{8})^2 = 1.6\%$ of the HRRR data. Using these pseudo-observations (pentagons in Fig.~\ref{subsample}, second row) in conjunction with our denoiser $D$, we generate an atmospheric state (plotted in the background) which is plausible and very similar to the ground truth full HRRR data (top row). This is similar to the exercise reported by \citeA{rozet2023scorebasedqg}, but for an operational weather model instead of a two-layer quasi-geostrophic model. We only perform the assimilation once for each set of (pseudo-)observations, rather than initializing with different noise inputs and producing an ensemble. This is to showcase what individual assimilated states look like, representing one sample from the posterior distribution. For the use as an ensemble technique, see section~\ref{assimobs}.

To test viability at the sparsity of actual weather station data, we next repeat the experiment on a more extremely subsampled data, to one in 18 or just 0.3\% of pixels (Fig.~\ref{subsample}, third row). Here, some of the smaller-scale features of the state are not reconstructed accurately, but the larger features are preserved, as should be expected. This still holds when we keep a similar number of pseudo-observations, but choose them in the same irregularly spaced configuration as the weather station data, leaving larger gaps between pseudo-observations (fourth row). We discuss the assimilation of actual observational data in sections~\ref{assimobs} and \ref{perfev}.

\begin{figure}[t]
\centering
\noindent\includegraphics[width=0.75\textwidth]{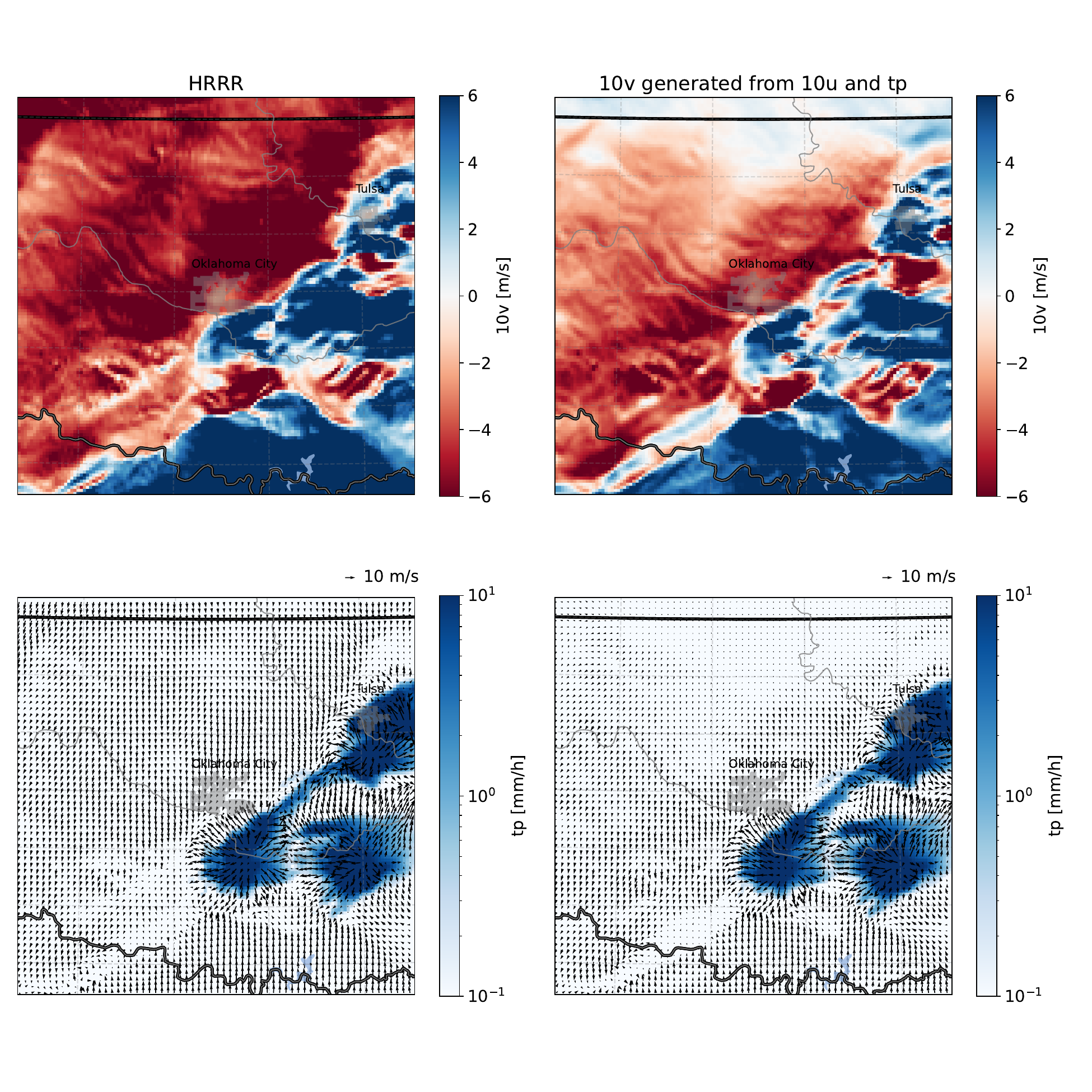}
\caption{Generating a left-out variable from other variables. We feed the model the HRRR 10u and precipitation, leaving out the 10v channel (top left). The model generates a reasonable 10v (top right). The bottom row shows HRRR tp overlaid with a quiver plot of 10u and 10v from HRRR (bottom left) and 10u and 10v from the model output. Note the wind arrows pointing away from the precipitation in both cases.}
\label{missingch}
\end{figure}

\subsection{DA of Missing Variables and Learned Physics}\label{sec:missing}

To further build confidence, we now test whether the model leverages physically appropriate multivariate relationships in the course of performing state estimation. Despite the fact that the model has been trained with and always outputs the same number of variables -- zonal wind, meridional wind, and precipitation rate -- we can leave out observations of one of these entirely and obtain a reconstruction purely based on the other variables. We show results of such an experiment (a single time step on 2017-05-28 03:00 UTC) in Figure~\ref{missingch}: We guide the generation with complete analysis data (dense grid of pseudo-observations)  of the zonal wind and total precipitation variables, leaving out all information about the meridional wind component. 

Encouragingly, the model generates a map of the held-out meridional wind that is qualitatively similar to the analysis-truth, in subregions where constraints are apparent. For instance, it succeeds at reconstructing the distinctive gust fronts associated with cold pools \cite{byers1949thunderstorm}: Evaporation of precipitation cooling the air, increasing its density and leading to downdrafts that diverge at the surface. Cold pools feed back on convection \cite{ross2004simple, feng2015mechanisms}, and are poorly represented in models, leading to errors in the representation of convection \cite{moncrieff2017simulation}. In our case, where there is precipitation in the analysis data, as well as diverging winds in the u-direction, the model reconstructs diverging winds in the v-direction. Here, the reconstruction agrees very well with the truth. We can hypothesize that even the change in large-scale wind direction from northerly in the northwestern to southerly in the southeastern half of the domain is inferred from the precipitation at the boundary, as this pattern is typical of a weather front. In the northern region of the domain, meridional wind is less constrained by precipitation, possibly causing a larger difference between model prediction and truth. The model associating strong precipitation with diverging near-surface winds provides evidence that it has learned physical behaviour from the training data. Given the importance of terrain for the onset of convection \cite{purdom1976some}, we would expect topography to additionally constrain precipitation. We leave an investigation of this effect, as well as a more quantitative study of cold pool dynamics, using e.g. wind gradients \cite{garg2020identifying}, for future work.

In sum, we find this qualitative evidence of learnt, physically valid multivariate relationships to add further confidence to the method's validity. The fact that these relations can be learned and usefully exploited across even within our limited set of three state variables should embolden future attempts that use considerably more ambitious state vectors, within which it is logical to assume additional physical relationships could be exploited in the state estimation. 

\subsection{Assimilation of Weather Station Observations}\label{assimobs}
\begin{figure}
\centering
\noindent\includegraphics[width=\textwidth]{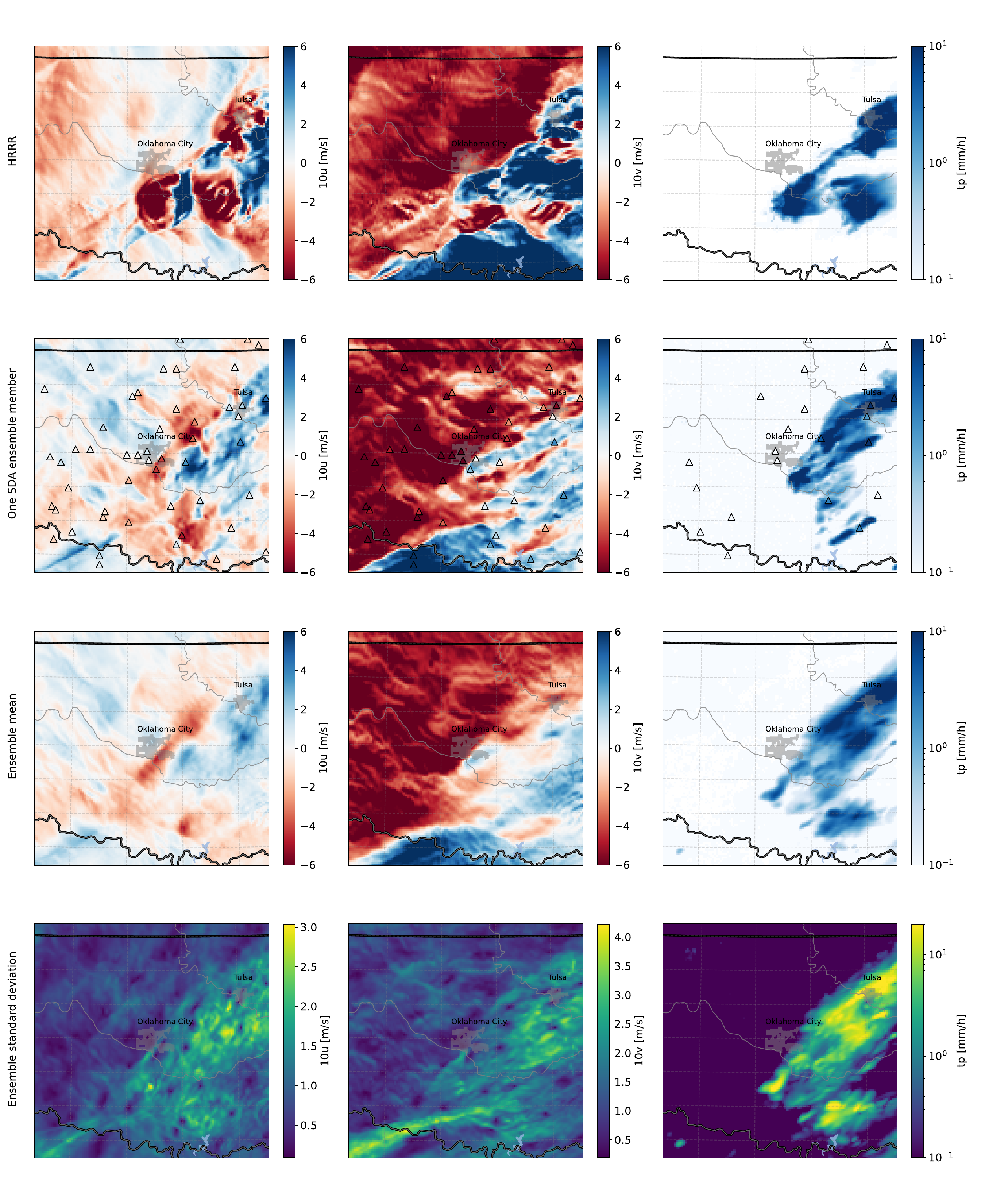}
\caption{SDA can produce stochastic ensembles of assimilated states. We assimilate the same station data as in Fig.~\ref{subsample}, but now generate a 20-member ensemble of states. We show the first member in the second row, the ensemble mean in the third, and the standard deviation in the fourth.}
\label{ensemble}
\end{figure}

Finally, we turn to assimilation of actual weather station data. Guiding the model's state estimation with the ISD observations provides equally physically plausible assimilated states as the pseudo-observations. We note that HRRR does not provide ground truth to compare the assimilated state from observations to. This is because HRRR in the station locations differs from the actual station observations, so even a perfect assimilation method would yield a state different from HRRR. Figure~\ref{subsample} shows one example (bottom row). The weather stations (triangles) are overlaid on the assimilated state. The stochasticity of the generation provides a natural way of producing ensembles of assimilated states, shown in Fig.~\ref{ensemble}: Even though the observational data and hyperparameters are the same, the state in the second row is different from the last row of Fig.~\ref{subsample}, but both agree in the observation locations. This becomes obvious in the standard deviation of the 20-member ensemble plotted in the bottom row of Fig.~\ref{ensemble}. Standard deviation is small around the locations where the assimilation is guided by observations, and encouragingly large in regions where internal variability from unpredictable dynamics exists - such as for convective precipitation in between station constraints in the east of the domain, and for meridional wind velocity associated with the imperfectly constrained convergence front to the southwest. This also implies that the choice of observation noise levels $\Sigma_P, \Sigma_u$ determines the diversity of ensemble states. These observation error statistics determine, according to equation~\ref{eq:sda-likelihood}, how strongly the reconstructed state should be constrained by observations. With a large value of the noise, the ensemble will have larger variance, as it is less constrained by the difference 
$\frac{\mathbf{y}-\mathcal{H}\left(\mathbf{\hat{x}}\right)}{\Sigma_y} $.

The mean, in the third row, shows a characteristic blurriness of ensemble means, resulting from averaging the individual members which disagree in small-scale features. 

\subsection{Performance Evaluation on Left-out Stations}
\label{perfev}
We quantitatively evaluate the model's performance on the full year of 2017 by comparing the assimilated state, which is obtained by guiding the model inference with a number of observations, with other, left-out observations. First, we test the dependence of SDA performance on the number of stations used for inference, then we fix the number of stations for inference and evaluation and study the performance of ensembles of assimilated states and compare to the HRRR analysis. 

\begin{figure}[h!]
\centering
\noindent\includegraphics[width=\textwidth]{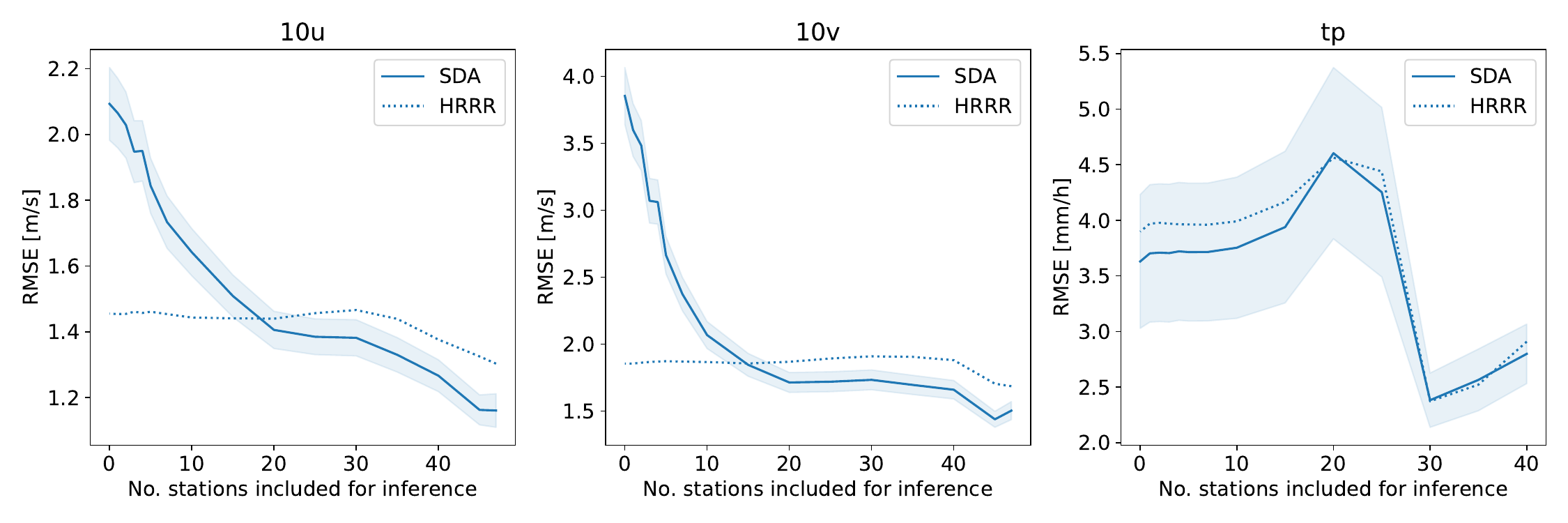}
\caption{Testing the dependence of assimilation on station density. Evaluating using data from the whole year of 2017, we vary the number of stations used for guiding the inference by the SDA framework. The resulting states are evaluated on the held-out stations, giving the RMSEs in each of the variables (solid lines). We also evaluate the RMSE of the HRRR analysis on the same held-out stations (dotted lines). }
\label{obssweep}
\end{figure}

\begin{figure}[h!]
\centering
\includegraphics[width=\textwidth]{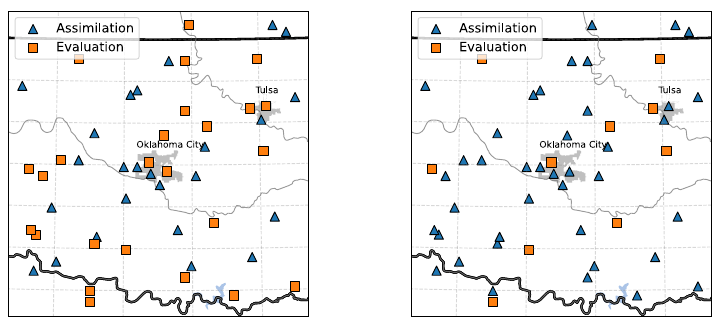}
\caption{For evaluation we leave out different numbers of randomly selected stations. For the results in Fig.~\ref{obssweep}, the left panel shows the 25 stations used for assimilation and inference each, which is where the wind RMSE of overtakes HRRR. For the results in Fig.~\ref{quanteval}, ten stations are left out for evaluation, as shown on the right.}
\label{statloc}
\end{figure}

The results indicate that, even in its crude, low-dimensional, prototype incarnation, SDA already provides more skillful estimates of surface wind than HRRR itself (noting that providing point-estimates is not the primary objective of the HRRR analysis). In Fig.~\ref{obssweep}, we show how the error of single SDA states on the evaluation stations decreases, the more station data we use for guiding the inference. We have 50 station locations in the region, so the number of evaluation stations can be found by subtracting inference stations from 50. The HRRR states do not depend on the number of stations used for inference or evaluation, so we expect them to be constant. Departures from a constant HRRR RMSE at large station number for inference can be explained by correspondingly small numbers of evaluation stations (small sample size), and consequently larger stochastic variability, impacting the estimate of RMSE. The intersections of the dotted and solid lines for the wind variables show that around 25 stations are enough for a single SDA assimilation to improve on the error of the HRRR analysis on the held-out stations. For a visualization of a split into 25 assimilation and 25 evaluation stations, see Fig.~\ref{statloc}. Note, that HRRR itself assimilates all of these stations. The results for precipitation are inconclusive, as the HRRR RMSE is not constant, implying large noise in precipitation RMSE at some stations. 

\begin{figure}
\centering
\includegraphics[width=\textwidth]{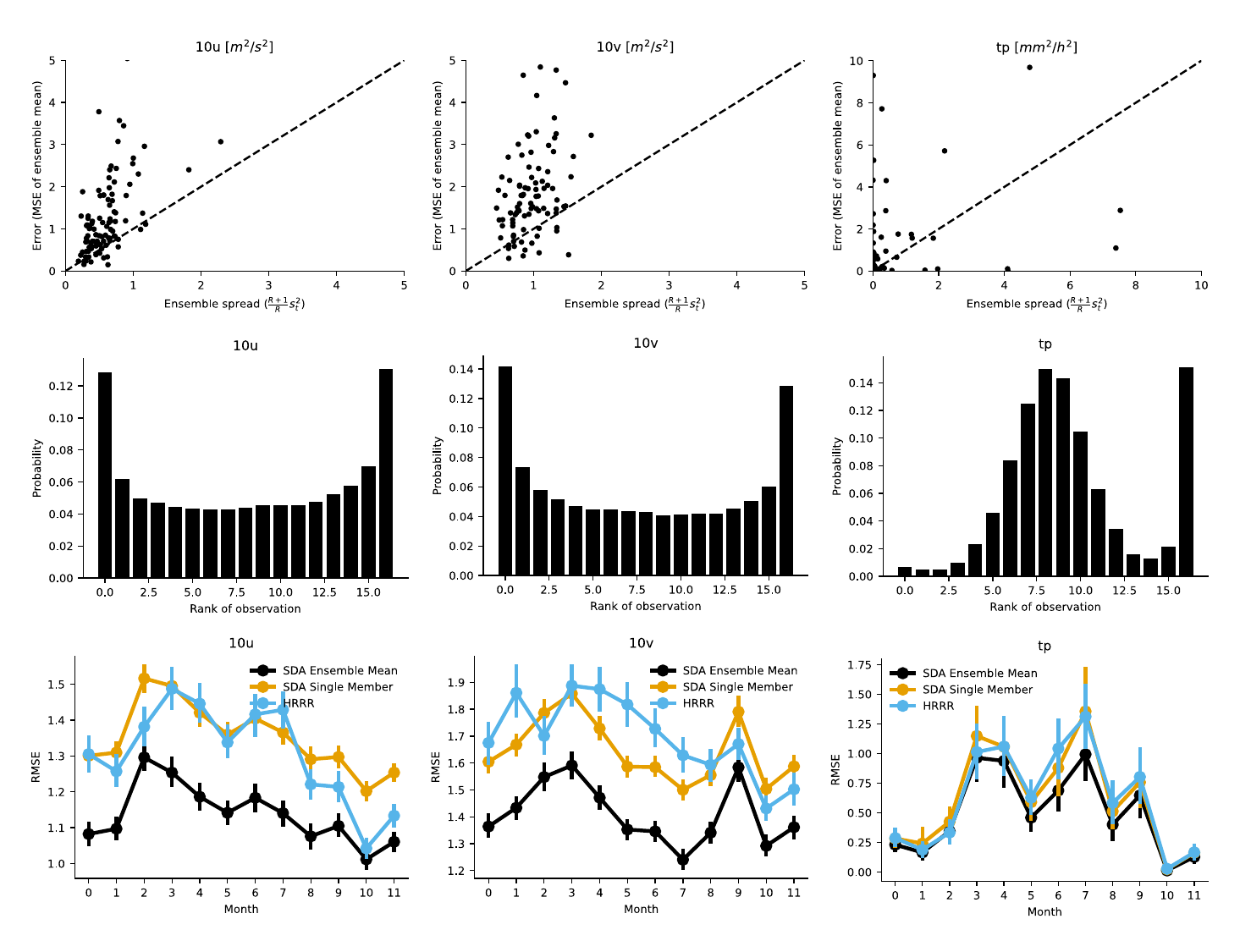}
\caption{
Quantitative evaluation of the SDA assimilation ensemble. 
The first row shows the spread-skill relationship of the evaluation observations.
To avoid overplotting, we only show 200 randomly selected samples of observed times/locations.
The second row  shows rank histograms of the 15-member ensemble
The third row shows the dependence of RMSEs on time of the year.
The error-bars in the RMSE panels are 95\% confidence interval of the mean value computed over bootstrapped samples.
}
\label{quanteval}
\end{figure}

\begin{table}
 \caption{Performance metrics of SDA ensembles and the HRRR analysis}
 \centering
 \begin{tabular}{l c c c c c c}
  \hline
    &\multicolumn{2}{c}{10u [m/s]}& \multicolumn{2}{c}{10v [m/s]} & \multicolumn{2}{c}{tp [mm/h]}  \\
    & SDA & HRRR & SDA & HRRR  & SDA & HRRR\\
 \hline
    CRPS  & 0.66 & — & 0.83 & — & 0.28 & — \\
    MSE\textsubscript{mean}  & \textbf{1.47}  & 2.08& \textbf{2.28}& 3.75& \textbf{4.39}& 5.82\\
    MSE\textsubscript{single}  & \textbf{2.00}  & 2.08& \textbf{2.99}& 3.75&6.79& \textbf{5.82}\\
   MAE\textsubscript{mean}  & \textbf{0.89} & 1.05& \textbf{1.13} & 1.38& \textbf{0.31}& 0.39\\
   MAE\textsubscript{single}  & \textbf{1.06} & 1.05& \textbf{1.30} & 1.38& \textbf{0.34}& 0.39\\
   Var\textsubscript{ens} &0.62 & — & 0.92& — &2.31 &—  \\

 \end{tabular}\label{tab:eval}
 \end{table}
Ensemble SDA appears to confirm competitive wind and precipitation state estimates across seasons (see Fig.~\ref{quanteval}). We fix the number of inference stations at 40 and evaluation stations at 10 (for a visualization, see Fig.~\ref{statloc}).  For a single-member ensemble of assimilated states, the performance is similar to the HRRR analysis. Moving to ensemble assimilation with 15 members, we show that we outperform deterministic HRRR analysis with around 10\% lower RMSEs.

Full results are given in Table~\ref{tab:eval}. Looking at the time-dependence of the performance, Fig.~\ref{quanteval}, bottom row, shows that while wind RMSEs are relatively constant in time,  precipitation RMSE increases in the summer period, which is characterized by deep convection and heavy rainfall. 

It is important to note that HRRR states provide a convenient but not ultimate baseline for km-scale state estimation in this comparison, as the HRRR DA scheme is designed with broader goals than pointwise matching to station data in mind. In particular, the HRRR DA is aiming to provide initial states for forecasts with a numerical model based on the Navier-Stokes equations. Nonetheless, HRRR is evaluated partially on its level of mismatch at station data scale \cite{james2022high}, and our main finding is that it is encouraging that a first attempt at km-scale SDA can match or exceed its performance on this metric.

We find ensembles from SDA to be under-dispersive and in need of further calibration for optimal utility. In ensemble forecasting, it is important that the ensemble spread is large where the prediction is uncertain. We also expect this in the assimilation context \cite{fortin2014should}. In Fig.~\ref{quanteval}, we show, for each variable, the spread-error diagram and rank histograms. For the spread-error diagram, we compare the ensemble variance (spread) to the MSE of the ensemble mean (error) of all time-samples. In Gaussian data assimilation, a well calibrated ensemble should lie around the dashed 1:1 line. We find that for all variables the spread is smaller than the observed error, meaning that the ensemble is underdispersive \cite{fortin2014should}. Rank histograms show the same effect: For each held-out observation, we determine how it ranks among the 15 ensemble members that are ordered by the predicted value. The u-shaped wind rank histograms show that in many cases, the entire ensemble is either predicting too-large or too-small values, i.e. the observations often are smaller or larger than all ensemble member predictions. For precipitation, the observations are often larger than the ensemble members, showing a low-bias in SDA precipitation states. We discuss how to improve the calibration in future work in the next section. 

\section{Conclusions}\label{sec:conclusions}

The weather simulation enterprise spans prediction and state estimation. For prediction, machine learning methods from the image and video generation community have already proved transformative. For state estimation, we have added to evidence that related methods in diffusion ML modeling for inverse problems also have encouraging potential.

In this study, we demonstrate, for the first time, score-based weather station data assimilation at km-scale. We show that - even in a crude, low-dimensional (three-variable) setup - state estimates of observations of surface wind fields are closer to held out station observations than HRRR analyses, with similar errors in precipitation (Note that all of the observations assimilated by the SDA were assimilated by the HRRR data assimilation system).  This is using only a few dozen observations and no remote sensing observations. We may expect extensions of regional SDA that incorporate these will enjoy additional skill gains. We further find evidence of learned physics, which our model can use to reconstruct missing variables. This means that we can effectively constrain unobserved variables in future data assimilation models that produce many more of the variables available in traditional numerical analyses. 

A key point is the simplicity of this tooling relative to traditional data assimilation methodology in numerical weather prediction. Crucially, SDA allows for very flexible addition of new observed data, as the station observations are only used in the inference phase. The training phase uses only HRRR analysis data. This is in contrast with the approach of \citeA{andrychowicz2023deep}, who use pairs of observations and analysis states in training. Different from traditional data assimilation approaches, we do not rely on a previous forecast from a computationally expensive numerical model, but implicitly use the distribution of possible weather states learned from years of analysis data. Our prior coming from an ML model leads to a large gain in speed, avoiding introduction of forecasting error. In this study, it also means that we do not use observations from previous time steps, even though this extension is possible in SDA \cite{rozet2023scorebased}.

While making a convincing case for SDA, we have also shown some limitations of a naive extension of the work of \cite{rozet2023scorebasedqg}, on the way to operational use. Firstly, we currently rely only on weather station observations, and only on a subset of the available measurements. We expect other observations like station observations of 2m temperature (important e.g. for fronts and cold pools), and pressure (more consistent winds) are relevant to constraining surface wind and precipitation. Nor have we used any remote sensing (e.g. dopper radar) data as all the gold standard mesoscale analysis products (MRMS, Stage IV, RTMA, HRRR) do. That said, fully exploiting these valuable data may require algorithmic improvements to both the training and inference. For example, their incorporation via a more complex observation operator $\mathcal{H}$ may be challenging: We tried to use an exponential in the observation operator for precipitation, rather than log-transforming the weather station data to match the distribution of data the diffusion model had been trained on. This led to numerical instability in the guided inference, presumably because initial noise states produced very large or even overflowing $\mathcal{H}(\mathbf{x}_\tau)$. 

Secondly, we show that our SDA ensembles are underdispersive. Calibration has not been previously reported for SDA methods, e.g. by \cite{qu2024deep, huang2024diffda}, and we show how it is essential to assess the method's performance. It is possible that some of the error reduction compared to observations could be simple variance reduction.
This could be related to mode collapse and low variances produced by our diffusion model, as discussed in Appendix~\ref{app:uncond}. It could be addressed by training the model on more (years or regions of) data, leading to better calibration of the unconditional diffusion (the prior).
It could also relate to our particular choice of hyperparameters (in particular noise parameters $\Sigma_P, \Sigma_u$), which were tuned for RMSE and CRPS, but not for ensemble dispersiveness. We also note that the modal approximation for the likelihood $ p(\mathbf{y} | \mathbf{x}_\tau)$ (eq. \ref{eq:sda-likelihood}) introduces an error, which may contribute to underdispersiveness of the ensemble. For future work it seems promising to evaluate this question by training a different model to directly approximate  $ p(\mathbf{x} | \mathbf{y})$, where $\mathbf{y}$ would be given by pseudo-observations taken from HRRR with modeled observational noise. Other approaches of modeling the likelihood could also be explored, and this is a topic of active research \cite{mardani2023variational, song2023pseudoinverse, rozet2024learning}. A benchmark dataset, comparing different methods, potentially on the simpler model of the quasi-geostrophic equations, would be of great benefit.

To use an assimilated state from traditional methods as an input to a (numerical) forecasting model, the state needs to be balanced so as to minimize artificial inertia-gravity waves in the model integration  \cite{peckham2016implementation}. With the three variables we have assimilated here, we cannot evaluate whether SDA would produce a balanced state, and to what extent digital filtering, i.e. removing imbalances for model initialization, would be necessary. We note that in samples that we calculated (surface) wind divergences in, we did not find unphysically large values.

Future research may also explore making use of the time-dimension of observational data, which would further constrain the atmospheric state. A more flexible approach that could assimilate data at arbitrary times would also improve assimilation, as some error is introduced by first interpolating station data to common points in time. 




%

\newpage
\appendix

\section{Denoiser in the SDA Framework}\label{app:edmsda}
In this project, we followed the EDM training framework of \citeA{karras2022elucidating}, which essentially trains a neural network that maps from a noisy state $\mathbf{x}_\tau$ and a noise level $\sigma_{\tau}$ to the denoised state $\mathbf{x}_0$. Meanwhile, in the the SDA framework, \citeA{rozet2023scorebased} train a network not to give the denoised state directly, but to output the noise pattern $\epsilon$ given the noisy state and the diffusion time $\tau$. This is why a model trained in the EDM framework cannot be directly used in SDA. However, such a diffusion model trained in the
EDM framework \(D\) can be translated into the more standard formulation of the SDA framework \(\epsilon_{\phi}(\mathbf{z},\tau)\) by using the relationship
\begin{equation}\label{denoiser}
\epsilon_{\phi}(\mathbf{z},\tau) = \left( \mu_{\tau}^{- 1}\mathbf{z} - D\left( \mu_{\tau}^{- 1}\mathbf{z};\mu_{\tau}^{- 1}\sigma_{\tau}^{s} \right) \right)\frac{\mu_{\tau}}{\sigma_{\tau}^{s}}
\end{equation}
where we call the noisy state \(\mathbf{z}\) (see below).

To derive this, we begin with some notation. Let $\mathbf{x}_0$ be a sample from the data, and then let $\mathbf{x}_\tau^{EDM}$ and $\mathbf{x}_\tau^{SDA}$ be the value of the EDM and SDA forward diffusion processes, respectively, at time $\tau$. Let $\epsilon_\tau$ be a Wiener process. The derivation consists of three parts. First we define the forward process and training target of EDM and SDA. Second, we must convert between the denoiser $D$ and the learned score function $\epsilon_\theta$ since SDA requires the latter. Finally, we relate $\mathbf{x}_\tau^{SDA}$ and $\mathbf{x}_\tau^{EDM}$.

In EDM, the forward process is defined by 
\begin{equation}\label{fwedm}
\mathbf{x}_\tau^{EDM} = \mathbf{x}_0 + \sigma_\tau \epsilon_\tau    
\end{equation}
and the denoiser is trained so that $D(\mathbf{x}_\tau^{EDM}, \tau)\approx \mathbf{x}_0$. 

In SDA, the foward process is defined by 
\begin{equation}\label{fwsda}
\mathbf{x}_\tau^{SDA} = \mu_\tau \mathbf{x}_0 + \sigma_\tau^s \epsilon_\tau    
\end{equation}
and the learned score function is trained so that $\epsilon_\phi(\mathbf{x}_\tau^{SDA}, \tau)\approx\epsilon_\tau$.

From this, we see that
\[
D(\mathbf{x}_\tau^{EDM}, \tau) + \sigma_\tau \epsilon_{\phi}(\mathbf{x}_\tau^{SDA},\tau) \approx \mathbf{x}_0 + \sigma_\tau \epsilon_\tau = \mathbf{x}_\tau^{EDM}
,\] and solving gives
\begin{equation}
    \epsilon_{\phi}(\mathbf{x}_\tau^{SDA},\tau) \approx \frac{\mathbf{x}_\tau^{EDM}-D(\mathbf{x}_\tau^{EDM}, \tau)}{\sigma_\tau}.\label{eq:d-score}
\end{equation}
While this derivation only shows an approximation relationship, it can be shown that this is exact from the ODE formulation of the backwards process.

It remains to express $\mathbf{x}_\tau^{EDM}$ in terms of $\mathbf{x}_\tau^{SDA}$, which is easily done by using the definitions of the forward process above (substituting $\mathbf{x}_0$ from \eqref{fwsda} into \eqref{fwedm})
\[
    \mathbf{x}_\tau^{EDM} = \mu_\tau^{-1} (\mathbf{x}_\tau^{SDA} - \sigma_\tau^s \epsilon_\tau) + \sigma_t \epsilon_\tau.
\]
We cancel the Wiener terms by setting $\sigma_\tau=\sigma_\tau^s/\mu^\tau$, as we are free to do, finally obtaining
\begin{equation}
    \mathbf{x}_\tau^{EDM} = \mu_\tau^{-1}\mathbf{x}_\tau^{SDA}. \label{xtedm}
\end{equation}
Substituting \eqref{xtedm} into \eqref{eq:d-score}, replacing $\mathbf{x}_\tau^{SDA}$ with the placeholder variable $\mathbf{z}$ completes the derivation.

\section{Hyperparameters}\label{app:hyperparams}
We perform minimal hyperparameter tuning on pseudo-observations from HRRR in 2017, in a similar setting as Fig.~\ref{subsample}. We find little dependence of RMSEs of our predicted states on number of denoising steps N. The sampling contains some steps of Langevin Monte Carlo corrections as discussed in section~\ref{sec:methods}, but similarly the increasing number C of corrections, and the correction size $\tilde{\tau}$ do not improve the results above the values reported in the second row of Table~\ref{tab:hyperparameters}. The size of the Langevin steps is given by $\delta = \tilde{\tau} \frac{dim(s)}{|| s||^2_2}$, so it is given by $\tilde{\tau}$ divided by the mean squared weight of the network $s$.  Reducing the value of $\Gamma$ to 0.001 from the 0.01 used by \citeA{rozet2023scorebased} was found to improve the results, particularly in the precipitation channel. For the observation error statistics $\sqrt{\Sigma_y}$, we evaluate the effect of using the values of standard deviations of observations with respect to HRRR data, as a heuristic for the noise of the observation. However we do not find improved skill with channel-wise observed values and therefore keep the original value of 0.1. The values of the second row of Table~\ref{tab:hyperparameters} are used throughout this study, except for section~\ref{sec:missing}, where the model diverges in the low-$\Gamma$ setting and we fall back to the original value of 0.01. Additionally we increase the numbers of steps and corrections of the sampling, as for this single sample computational efficiency is less important. We use the same noise schedule for inference as \citeA{rozet2023scorebased}, namely
\[
\mu_{\tau} = \cos(\omega \tau), 
\omega = \cos^{- 1}\sqrt{10^{- 3}}
, \sigma_{\tau}^{s} = \sqrt{1 - \mu_{\tau}^{2}}.
\]
\begin{table}[t]
 \caption{Hyperparameters used for experimental results}
 \centering
 \begin{tabular}{l c c c c c }
  \hline
    & N & C & $\tilde{\tau}$ & $\sqrt{\Sigma_y}$  & $\Gamma$ \\
 \hline
 missing channel  & 256 & 10 & 0.3 & 0.1 & 0.01 \\
 all other results  & 64 & 2 & 0.3 & 0.1 & 0.001 \\
 \hline
 \end{tabular}\label{tab:hyperparameters}
 \end{table}

\section{The climate of the diffusion model}\label{app:uncond}

\begin{figure}[htbp]
\includegraphics[width=\textwidth]{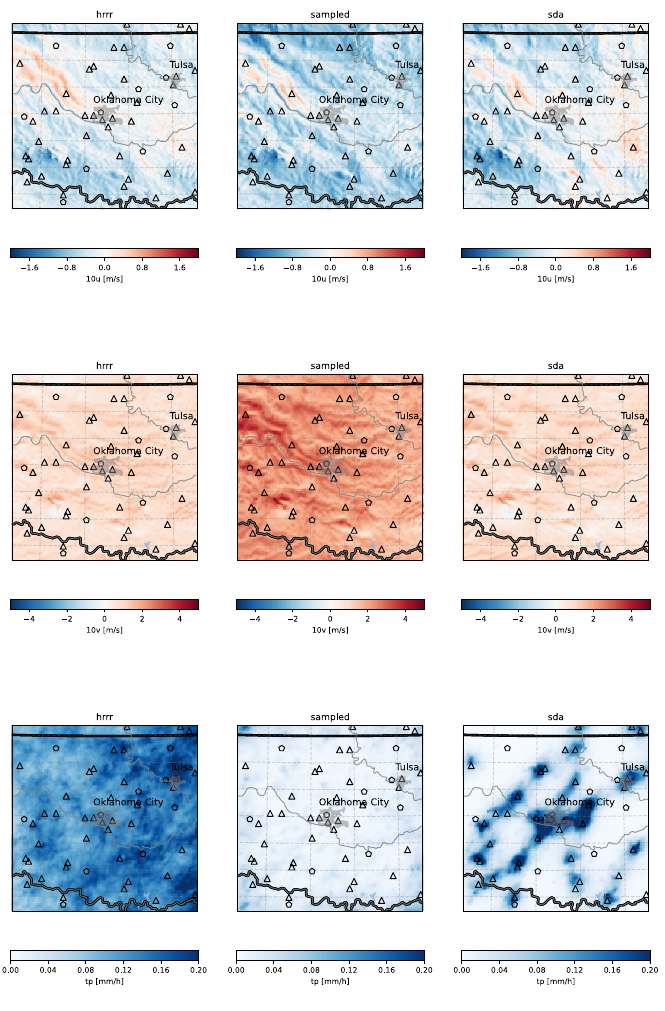}
\caption{Time mean maps for the three channels in the samples from HRRR, for the unconditional samples from the generative model (sampled), and those with guidance from station data (SDA).  \label{fig:allthree}}
\end{figure}

\begin{figure}
\centering
\includegraphics{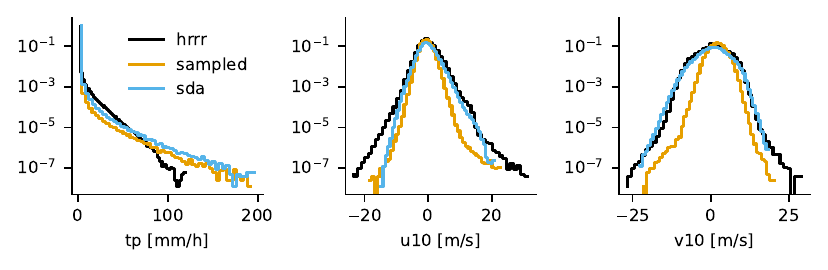}
\caption{The same data as shown in Figure \ref{fig:allthree}, here showing histograms of the respective distributions.\label{fig:climate-histo}}
\end{figure}

In this section, we examine the quality of the unconditional diffusion prior from a statistical perspective. We compare the unconditional generation samples of one year's worth of data with those from HRRR and SDA assimilated states of 2017, our test year. A map of the time-mean for all channels is shown in Fig.~\ref{fig:allthree}. While it seems in the time mean spatial structures are mostly well represented, showing signs of learned topography, the unconditional samples have some bias towards higher values of 10v and too little precipitation. The 10v bias is corrected by incorporating observations (SDA), but in precipitation SDA shows increases exclusively over the stations used for assimilation (triangle). This make some sense since the model tends towards too little precipitation on average, so the assimilation only ever nudges in the positive direction. A similar picture is apparent in the histograms shown in Fig.~\ref{fig:climate-histo}.  While for the winds, guiding with observations brings the distributions closer, for precipitation the model keeps underestimating normal rain amounts and occasionally predicts unphysically large values. We attribute this to the use of the exponential transform to go from model outputs to precipitation. In future work, approaches for modeling heavy-tailed distributions in diffusion models by \cite{pandey2024heavy} could be explored. 

\section{Skill metrics}
We evaluate the skill of single member and ensemble SDA against HRRR using the metrics Mean Squared Error (MSE) or its root (RMSE), Mean Absolute Error (MAE), and Continuous Ranked Probability Score (CRPS). MSE is calculated here from the error $\varepsilon_t$ of an individual (timestamp-) state $t$ as 
\begin{equation}
    \varepsilon^t = \mathbf{y}_E^t-\mathcal{H}_E(\mathbf{\bar{x}}^t)
\end{equation}
where $\mathbf{y}_E^t$ is the subset of observations for evaluation at locations $E$, $\mathbf{\bar{x}}^t$ is the mean of the SDA ensemble, and $\mathcal{H}_E$ is the observation operator for the evaluation locations. In the case of HRRR data, instead of $\mathbf{\bar{x}}^t$, we just use the single HRRR state $\mathbf{x}_{HRRR}^t$. Given $T$ timestamps for evaluation, MSE is just 
\begin{equation}
    MSE = \frac{1}{T}\sum_{t=0}^T (\varepsilon^t)^2.
\label{eq:mse}
\end{equation}
Likewise, MAE is 
\begin{equation}
    MAE = \frac{1}{T}\sum_{t=0}^T |\varepsilon^t|
\end{equation}

For the probabilistic metrics, we need to include the assumed observation noise model in the metrics. Let, the pseudo-observations be defined by $\tilde{\mathbf{y}}^t_r \sim p(\cdot|\mathbf{x}_{r}^t)$ independent for $r=1,\ldots, R$ where $R$ is the ensemble size. Here $p(\cdot|\mathbf{x}_{r}^t)$ stands for the specific observation operators defined in \ref{sec:ISD}. The observational noise is also added to the ensemble members to find the rank of the observation in the rank histograms.

Then, CRPS for a single time $t$ is defined following [equation eFAIR]\cite{zamo2018estimation}
\begin{equation}
    CRPS^t = \frac{1}{R} \sum_{r=1}^R |\tilde{\mathbf{y}}_{r}^t - \mathbf{y}^t_E | +\frac{1}{2R(R-1)}\sum_{r,q =1}^R |\tilde{\mathbf{y}}^t_r  - \tilde{\mathbf{y}}^t_q |
\end{equation} 
where $r, q$ denote the ensemble members.
Then, the overall CRPS is obtained by averaging over times $t$. To evaluate the model calibration, we compare the MSE with the ensemble spread, following \cite{fortin2014should} given by the bias-corrected time-average of the single-timestep variances of the ensemble $s_t^2$, such that
\begin{equation}
    \left(\frac{R+1}{R}\right) \frac{1}{T } \sum_{t=0}^T s_t^2; \ \
    s_t^2 = \frac{1}{R-1} \sum_{r=1}^R  \left( \tilde{\mathbf{y}}^t_r  - \overline{\tilde{\mathbf{y}}^t}   \right)^2.
\end{equation}
where the factor $\frac{R+1}{R}$ is to give an unbiased estimate with a small number (in our case 15) of ensemble members.

\section{Notation}\label{app:notation}
We use here the data assimilation notation following \citeA{ide1997unified}, rather than that of \citeA{rozet2023scorebased} in order to be more accessible for the data assimilation community. This just means that for the observation operator, we use $\mathcal{H}$ instead of $\mathcal{A}$, and for the observation error covariance matrix we use $\mathbf{R}$ instead of $\Sigma_y$, with $\Sigma_y$ used for its diagonal entries $\Sigma_{P}$, $\Sigma_{\mathbf{u}}$. Further, diffusion time is noted ${\tau}$ instead of $t$ to avoid confusion with physical time which plays no role in the assimilation as we work with snapshots, and is only used in the testing on different (independent) timestamps. Lastly, our states and observations are noted $\mathbf{x}$ and $\mathbf{y}$, understood as vectors. 

\section{Open Research}
The data used here is publicly available from the National Oceanic and Atmospheric Administration (NOAA). NOAA ISD data \cite{noaaisd} can be obtained freely from the NOAA website, with a useful search functionality under \url{https://www.ncei.noaa.gov/access/search/data-search/global-hourly}.
HRRR \cite{hrrr} data can be obtained freely from NOAA under \url{https://home.chpc.utah.edu/~u0553130/Brian_Blaylock/cgi-bin/hrrr_download.cgi}. 
Code for the EDM framework \cite{karras2022elucidating} is publicly available under \url{https://github.com/NVlabs/edm}, and for SDA \cite{rozet2023scorebased} under \url{https://github.com/francois-rozet/sda}. 
The code used for preprocessing ISD and applying SDA as shown in this paper as well as the data used for the figures together with figure scripts has been incorporated into the NVIDIA/PhysicsNeMo repository, which has been archived in its state at the time of publication under \url{https://doi.org/10.5281/zenodo.15083507} \cite{physicsnemo_contributors_2025_15083508}.  The code for this project can be found in the archived PhysicsNeMo repository by navigating to \texttt{examples/weather/regen/}.


\acknowledgments
This research used resources of the National Energy Research Scientific Computing Center (NERSC), a Department of Energy Office of Science User Facility using NERSC award ERCAP0028849. We would like to thank Dale Durran for valuable discussions and feedback on the project, in particular on how to relate HRRR to observations. We also commend François Rozet for his very clear and reproducible SDA code. We thank three anonymous reviewers for their helpful comments. 



%
\newpage
\bibliography{references} 




%
%
%
%
%

\end{document}